# Generating Dataset For Large-scale 3D Facial Emotion Recognition

**Faizan Farooq Khan**[*]
International Institute Of Information Technology, Hyderabad
Hyderabad, India
`faizan.farooq@students.iiit.ac.in`

**Dr. Zulqarnain Gilani**
Edith Cowan Univeristy
Joondalup WA 6027, Australia
`s.gilani@ecu.edu.au`

## Abstract

The tremendous development in deep learning has led facial expression recognition (FER) to receive much attention in the past few years. Although 3D FER has an inherent edge over its 2D counterpart, work on 2D images has dominated the field. The main reason for the slow development of 3D FER is the unavailability of large training and large test datasets. Recognition accuracies have already saturated on existing 3D emotion recognition datasets due to their small gallery sizes. Unlike 2D photographs, 3D facial scans are not easy to collect, causing a bottleneck in the development of deep 3D FER networks and datasets. In this work, we propose a method for generating a large dataset of 3D faces with labeled emotions. We also develop a deep convolutional neural network(CNN) for 3D FER trained on 624,000 3D facial scans. The test data comprises 208,000 3D facial scans.

## 1 Introduction

Emotional expressions are the behaviors that communicate our emotional state or attitude to others. They have always been used and form an essential part of interpersonal communication. Emotions are expressed in different forms, often missed by the naked eye. Recently deep learning has become the mainstream method to promote FER. FER has gained traction in the past few years, and reliable FER systems are required in affect-aware machines and devices. Such systems can understand human emotion and interact with users more naturally. Most of the work has been done in the 2D modality, resulting in a significant gap between 2D and 3D FER. The lack of availability of a large 3D data set is responsible for the slow development of the FER in 3D modality.

The 3D facial scans have inherent advantages over 2D images. The 2D images suffer from variations of head-pose, illumination, occlusion, and identity, whereas 3D facial scans are uninfluenced by these factors (1). The ability to capture all the muscle movement accurately, regardless of the lighting and pose variations, by using 3D instruments provides high resolution 3D facial scans where the muscle activities that change with expressions are visually evident, which is beneficial for FER.

In this paper, we present a technique for data augmentation that introduces non-linear heterogeneous variations in 3D shape and facial expressions to generate a dataset of 832K 3D scans of 100 unique identities. Existing 3D datasets for FER are several orders of magnitude lower than ours, as shown in Table 1.

Apart from data, the expression recognition algorithm itself is a critical component. Using networks trained on 2D images to perform 3D face recognition is simplistic and sub-optimal as 3D data has its peculiarities defined by the underlying shape and geometry. To the best of our knowledge, there is no deep network trained on a dataset comparable to ours in size, which is designed specifically for 3D

---

[*]Use footnote for providing further information about author (webpage, alternative address)—*not* for acknowledging funding agencies.

Table 1: Details of 3D Datasets for FER

| Dataset | Expressions | Identities | Facial Scans |
|---|---|---|---|
| BU3DFE (19) | 7 | 100 | 2500 |
| BU4DFE (20) | 6 | 6 | 60,600 |
| FRGC v.2 (22) | 6 | 466 | 4007 |
| Bosphous (21) | 7 | 105 | 4652 |
| **Ours** | 7 | 100 | 832,000 |

FER. We cover this research gap and propose a Deep 3D expression recognition network suited for 3D face data and trained from scratch on 624K 3D faces.

In a nutshell, our contributions are as follows: (1) Large scale dataset: We present a method for generating a large corpus of labeled 3D facial expression data for training CNNs. Our dataset contains 832K 3D facial scans of 100 identities, each consisting of 6 basic expressions and a neutral face, highly rich in shape variations. (2) Deep 3D Facial Emotion Recognition Network. We propose a deep CNN explicitly designed for 3D FER and trained on 832k 3D faces. We show its effectiveness by achieving an end-to-end Rank-1 recognition rate of 88% on the BU3DFE dataset.

## 2 Related Works

Facial Expression recognition has been vastly studied, but the studies have focused majorly on 2D data for the most part. 3D FER has recently started to become an extensive research field, and some early attempts include (2; 3; 4; 5; 6), and the most recent (7; 8; 9). There have been efforts made (8; 11) to try methods that tend to both 2D and 3D modalities.

### 2.1 Manually-Crafted Feature-Based Method

The use of curvature-based features from 3D facial models to detect areas with high curvatures, such as the nose tip and eye cavities is followed in (12; 13). A segmentation-based approach using K-means to discard the background and locate the candidate faces using edge and ellipsoid detection is used in (14). Random Forests are used by (15), each pixel is assigned a body part label, including the face. The approach used by (15) is extended in (16) which uses Graph Cuts to optimize the Random Forest probabilities.

### 2.2 Deep Learning based 3D Facial Expression Recognition

Deep CNN's for FER of six basic expressions is proposed in (17). The CNN's are trained on the 2D facial appearance and the 3D face shape, respectively, obtained from the BU-3DFE database. The use of hand-crafted descriptors in capturing the local shape and texture information is introduced in (18); this work is extended by (8), where deep learning techniques are introduced in which they achieve state-of-the-art performances on the BU-3DFE database at that time.

### 2.3 Multi Modal Techniques

The use of 2D texture information incorporated along with 3D features is done in (23). After learning the SVM models from the 2D and 3D data separately, classification is performed. In 2015, (9), proposed to fuse the key features obtained from the geometric and textured domains, to investigate how the overall performance is affected. They conducted experiments on the BU-3DFE database, demonstrating the effectiveness of combing texture and depth cues.

## 3 Data Generation Process

We use 3D facial scans of 100 individuals from the BU3DFE dataset to train and test our deep network. For an individual, we have 6 basic expressions of 4 levels and a neutral face, which gives us 25 facial scans for each identity. Out of all 25 facial scans, we make use of level 2, level 3, and the neutral expression, giving us 13 facial scans for each identity. Inspired by the work done by (29), we establish dense correspondence over 15K 3D vertices on the faces from this dataset, using the keypoints-based algorithm. The goal now is to grow the dataset by generating faces from the space spanned by the trio of densely corresponding real 3D faces of distinct identities. We make use of the idea put forward by (34) and expand on it for facial expressions. To ensure that the identities in the trio are as "distinct" as possible, we select the face trio with the maximum non-rigid shape difference and the trio having the same expression to ensure the generated face has the same expression as that of original faces.

Table 2: Comparison of emotion recognition accuracy (%) on BU3DFE AND Bosphorus dataset.

| Method | Dataset | Expressions | Train Test Ratio | Modality | Accuracy |
| --- | --- | --- | --- | --- | --- |
| Li et al. (30) | BU3DFE | 6E | 60:40 | 2D+3D | 86.86% |
| Jan et al. (9) | BU3DFE | 6E | 60:40 | 2D+3D | 90.04% |
| Sheng et al. (32) | BU3DFE | 6E | 83:17 | 3D | 92.1% |
| Khashman et al. (33) | BU3DFE | 7E | 80:20 | 3D | 95.2% |
| **Ours**(Kernel=7) | BU3DFE | 7E | 60:40 | 3D | 88% |
| **Ours**(Kernel=5) | BU3DFE | 7E | 60:40 | 3D | 86.7% |
| **Ours**(Kernel=3) | BU3DFE | 7E | 60:40 | 3D | 86% |
| Azazi et al. (10) | Bosphorus | 7E | 90:10 | 3D | 79% |
| Hariri et al. (31) | Bosphorus | 7E | 90:10 | 3D | 86.17% |
| Zhang et al. (24) | Bosphorus | 6E | 80:20 | 3D | 92.2% |
| **Ours**(Kernel=7) | Bosphorus | 7E | 0:100 | 3D | 71.4% |
| **Ours**(Kernel=5) | Bosphorus | 7E | 0:100 | 3D | 70.8% |
| **Ours**(Kernel=3) | Bosphorus | 7E | 0:100 | 3D | 70.4% |

Let the faces be represented by $F_i = [x_p, y_p, x_p]^T$, where $i = 1, ..., N, p = 1, ..., P; N = 100$ and $P = 15,000$. The shape difference between faces $F_i$, $F_j$ and $F_k$ is defined as

$$D(i,j,k) = \frac{\gamma_{ij} + \gamma_{ji} + \gamma_{ik} + \gamma_{ki} + \gamma_{jk} + \gamma_{kj}}{6} \quad (1)$$

where, $\gamma_{ij}$ is the amount of bending energy required to deform 3D face $F_i$ to face $F_j$. Extending the 2D thin-plate spline model (10) to our case, we calculate the bending energy as, $\gamma_{ij} = x^T B x + y^T B y + z^T B z$ where $x, y$ and $z$ are the vectors containing the $x, y$ and $z$ coordinates of $P$ points in face $F_j$ and $B$ is the bending matrix, which is defined as the $P \times P$ upper left matrix of

$$\begin{bmatrix} K & S \\ S^T & 0 \end{bmatrix}^{-1}$$

Here, $K(a,b) = ||F_i^a - F_i^b||^2 \, log||F_i^a - F_i^b||$ with $a, b = 1, ..., P, S = [1, x^j, y^j, z^j]$, and 0 is a $P \times 4$ matrix of zeros. We select 832,000 trios of 3D faces with maximum shape difference D(i, j) from the possible 2,102,100 trios. The possible trios are calculated as $\binom{100}{3}$ for each expression, times the number of expression, i.e, 13. Since the 3D faces in each trio are in dense correspondence to each other, a new face **F** is generated from the linear space of each trio (i, j, k) as $\frac{[x_i^p, y_i^p, z_i^p] + [x_j^p, y_j^p, z_j^p] + [x_k^p, y_k^p, z_k^p]}{3}$. The 3D pointcloud generated from each trio is then used to generate a three channel image following the steps by (34). The three channels consist of depth, azimuth and elevation. The depth channel is generated by fitting a surface of the form z(x, y) to the 3D pointcloud using the gridfit algorithm (13). The surface normals of the original point-cloud are calculated in spherical coordinates ($\theta, \varphi$) where $\theta, \varphi$ are the azimuth and elevation angles of the normal vector. The surfaces of the form $\theta(x, y)$ and $\varphi(x, y)$ are fitted to the azimuth and elevation angles using a similar x, y grid to the depth channel to make the second and third channels. The three channels are normalized on the 0-255 range and can be rendered as an RGB image. This image is passed through a landmark identification network (18) to detect the nosetip. With the face centered at the nosetip, we crop a square of 224 × 224 pixels.

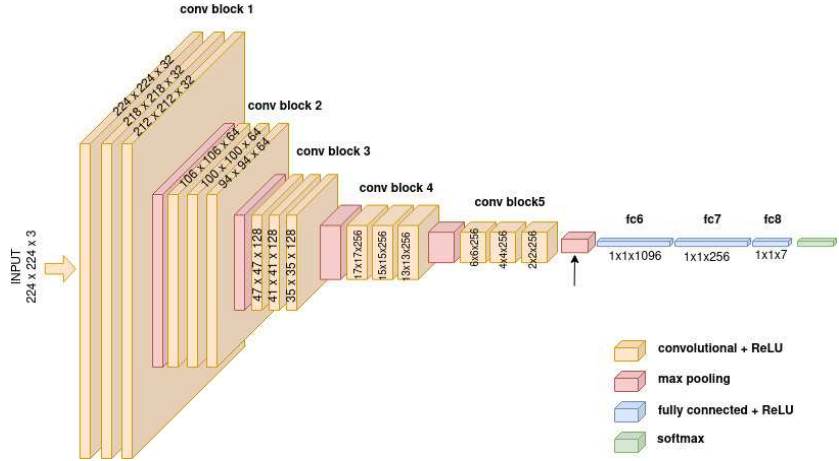

Figure 1: Architecture of our proposed network.

## 4 Network Architecture

The success of the recent deep network in 2D FER inspired us to propose a deep convolutional neural network that is suited to 3D data. 2D images exhibit significant texture variations over small regions and (35) network was designed for such case. In contrast, 3D facial surfaces are generally smooth, and hence filters with larger kernel sizes would better suit this type of data. This claim is strengthened by the results shown in Table 2.

The skeleton architecture of our network follows (35) but with a change in the convolution layers (see Figure 1 for details). We aim to minimize the average prediction binary cross-entropy loss after the softmax layer by learning the parameters of a network designed to classify the six basic emotions and the neutral face. After the network is trained, dropout layers are removed before testing.

## 5 Limitations

The limitations of our work can be divided into two parts. First, the synthetic dataset is generated using only 100 unique identities. Although the synthetic faces were generated from the space spanned by a trio of densely corresponding real 3D faces of distinct identities, the low number of identities lead to the generation of faces that are similar to other faces. This claim is further strengthened when the network is tested on the unseen test set. The set consists of identities not used during the training process. Training on similar faces prevents the generalization of the network leading to poor results. Second, the different available 3D datasets have been collected in controlled environments with every dataset captured by different instruments. The scans captured from different types of equipment result in different properties between two datasets, making it harder for a network trained on one dataset to perform on the other dataset.

## 6 Conclusion

Our work aims to bridge the vast gap between research advancements in 2D and 3D FER algorithms especially in the context of deep learning. The results are not able to beat the state-of-the-art algorithms but show promise by achieving 71.4% accuracy on the complete Bosphporus dataset which is not used for training. A combination of 3D datasets can provide more variation with a greater number of identities, the dataset generation technique described in the paper can yield better results in this case. In this work, we also show the difference between 2D and 3D facial data and how larger kernels perform better on 3D data while the case is the opposite for 2D datasets.